\documentclass[letterpaper]{article}
\usepackage{aaai2026}
\usepackage{times}
\usepackage{helvet}
\usepackage{courier}
\usepackage[hyphens]{url}
\usepackage{graphicx}
\usepackage{graphicx}
\usepackage{natbib}
\usepackage{caption}
\usepackage{algorithm}

\usepackage{algpseudocode} 
\algtext*{EndIf}
\algtext*{EndFor}
\usepackage{enumitem}
\usepackage{amsmath}
\usepackage{amsthm}
\usepackage{amssymb}
\usepackage{xcolor}
\usepackage{tikz}
\usepackage{booktabs}
\usepackage{multirow}
\usepackage{arydshln}
\usepackage{pgfplots}
\pgfplotsset{compat=1.18}
\usetikzlibrary{calc, positioning, arrows, shapes, intersections}
\usepackage[pro]{fontawesome5}
\newtheorem{example}{Example}
\newtheorem{definition}{Definition}
\newtheorem{proposition}{Proposition}
\usepackage{pgfplots}
\usepgfplotslibrary{fillbetween}

\definecolor{col1}{rgb}{0.283187, 0.125848, 0.44496}
\definecolor{textcol1}{gray}{1}
\definecolor{col2}{rgb}{0.132268, 0.655014, 0.519661}
\definecolor{textcol2}{gray}{0}
\definecolor{col3}{rgb}{0.83527, 0.886029, 0.102646}
\definecolor{textcol3}{gray}{0}

\definecolor{a4}{rgb}{0.13, 0.53, 0.20}
\definecolor{a3}{rgb}{0.40, 0.80, 0.93}
\definecolor{a2}{rgb}{0.27, 0.43, 0.67}

\newcommand{\agentDelayed}{\agent_1}
\newcommand{\agentWithFlexibility}{\agent_2}
\newcommand{\agentNoDelay}{\agent_3}
\newcommand{\agentOther}{\overline{\agent}}

\newcommand{\states}{S}
\newcommand{\SIPPedges}{E}
\newcommand{\edgeDurationVar}{\delta}
\newcommand{\edgeDuration}[1]{\edgeDurationVar(#1)}
\newcommand{\state}{s}
\newcommand{\statedef}{\state = \langle \configuration, \interval \rangle \in \states}
\newcommand{\configuration}{x}
\newcommand{\interval}{i}
\newcommand{\edge}{e}
\newcommand{\edgedef}{\edge = \langle \stateOne, \stateTwo, \interval \rangle}
\newcommand{\stateOne}{\state_1}
\newcommand{\stateTwo}{\state_2}
\newcommand{\nodeOne}{u}
\newcommand{\nodeTwo}{v}
\newcommand{\nodeThree}{z}
\newcommand{\nodeCritical}{w}
\newcommand{\timeVar}{t}
\newcommand{\timeStart}{\timeVar_s}
\newcommand{\timeEnd}{\timeVar_e}
\newcommand{\intervaldef}{\interval = [\timeStart, \timeEnd \rangle}
\newcommand{\problemStartTime}{\timeVar_o}
\newcommand{\startConfig}[1]{\configuration_o^{#1}}
\newcommand{\goalConfig}[1]{\configuration_g^{#1}}

\newcommand{\startTime}[1]{\timeVar_o^{#1}}
\newcommand{\tmpFlex}{m}
\newcommand{\waitTimeSteps}[2]{\tau^{#1}_{#2}}
\newcommand{\delay}{d}
\newcommand{\delayFunc}[1]{{#1}^{\mathrm{\delay}}}
\newcommand{\flexFunc}[1]{{#1}^\mathrm{F}}

\newcommand{\earliestWaitTime}{\zeta}
\newcommand{\earliestStartTime}{\alpha}
\newcommand{\unsafeTime}{\beta}

\newcommand{\transitTime}{\Delta}

\newcommand{\atfFunc}{A}
\newcommand{\pathParam}{p}

\newcommand{\maeder}{\langle \network, \agents, \chars \rangle}
\newcommand{\maedersol}{\mathcal{M}}
\newcommand{\network}{N}
\newcommand{\agents}{\mathcal{A}}
\newcommand{\agent}{a}
\newcommand{\chars}{\mathcal{C}}

\newcommand{\flexVar}{f}
\newcommand{\flex}[2]{\flexVar_{#1}(#2)}
\newcommand{\maxTime}{\timeVar^\mathrm{max}}
\newcommand{\sol}{\Pi}
\newcommand{\plan}[1]{\pi_{#1}}
\newcommand{\planLength}[1]{n_{#1}}
\newcommand{\planLengthMax}{n}
\newcommand{\numAgents}{k}
\newcommand{\stateIndex}{j}
\newcommand{\problem}{Flexible Multi-Agent Delay Replanning}
\newcommand{\tippingTime}{\timeVar^\circ}
\newcommand{\tippingLoc}{\configuration^\circ}
\newcommand{\tipping}{(\tippingLoc,\tippingTime)}
\newcommand{\IntroducedFlexibility}{\texttt{\flexVar}}

\title{Precomputing Multi-Agent Path Replanning Using Temporal Flexibility}

\author{
    Issa Hanou\textsuperscript{\rm 1},
    Eric Kemmeren\textsuperscript{\rm 1},
    Devin Wild Thomas\textsuperscript{\rm 2},
    Mathijs de Weerdt\textsuperscript{\rm 1}
}
\affiliations{
    \textsuperscript{\rm 1}Delft University of Technology\\
    \textsuperscript{\rm 2}University of New Hampshire\\
    \{i.k.hanou, m.m.deweerdt\}@tudelft.nl
}

\begin{document}

\maketitle

\begin{abstract}
    Executing a multi-agent plan can be challenging when an agent is delayed, because this typically creates conflicts with other agents.
    So, we need to quickly find a new safe plan. 
    Replanning only the delayed agent often does not yield an efficient plan, and sometimes cannot even yield a feasible one. 
    On the other hand, replanning other agents may lead to a cascade of changes and delays, and it is computationally expensive. 
    We show how to efficiently replan a single delayed agent by tracking and using the \emph{temporal flexibility} of other agents while avoiding cascading delays. 
    This flexibility is the maximum delay that the agent can take without changing the order with agents other than the initially delayed agent, or further delaying other agents.
    Our algorithm, FlexSIPP, precomputes all possible plans for the delayed agent and returns the changes to the other agents within the given scenario.
    We demonstrate our method in a real-world case study of replanning trains in the densely-used Dutch railway network and in the MovingAI MAPF benchmark set.
    Our experiments show that FlexSIPP provides effective solutions relevant to real-world adjustments, and within a reasonable timeframe.
\end{abstract}

\begin{links}
    \link{Code}{https://doi.org/10.5281/zenodo.23187201}
\end{links}

\section{Introduction}
\label{sec:intro}
When executing a plan in a multi-agent environment, delays frequently occur, leading to conflicts and disrupting the plan's previously safe execution. 
Resolving these conflicts requires deciding which agents' plans to disrupt and by how much.
As conflicts compromise the safe execution, a decision must be made swiftly, while replanning takes time.

In this paper, we address the multi-agent path replanning problem, where a single agent needs to be replanned (e.g., due to a delay), and our objective is to recover a safe plan as quickly as possible.
Full replanning of all agents can be computationally expensive, so we consider the subset of these problems where other agents may only be delayed, not rerouted or re-ordered, and this delay must not introduce additional conflicts.
These constraints are inspired by the railway network application.
We quantify an agent's ability to absorb a delay as its \emph{temporal flexibility}, which is the maximum delay it can take without changing the order in which other agents use a resource or further delaying them.
The temporal flexibility is easily precomputed from a given plan.
This allows us to identify the set of \emph{tipping points} for the delayed agent, which are the time points until which the delayed agent can use another agent's temporal flexibility and thus go first, and after this tipping point, the other agent should go first, and the ordering of the agents thus changes.
FlexSIPP also provides a second use case: given a single delayed start time, return a plan minimizing the total delay.

\begin{figure}[t]
    \centering
    \begin{tikzpicture}
        \def\cellsize{0.65cm}
        \def\iconscale{1}
        \def\width{7}
        \def\height{2}

        \draw[dotted, thick] (0,1*\cellsize) -- (\cellsize+\width*\cellsize,1*\cellsize);
        \draw[dotted, thick] (0,2*\cellsize) -- (\cellsize+\width*\cellsize,2*\cellsize);

        \foreach \i in {1,...,\width} {
            \draw[dotted, thick] (\i*\cellsize, 0) -- (\i*\cellsize, 3*\cellsize);
        }
        \draw[thick] (0,0) rectangle ++(\cellsize+\width*\cellsize,\cellsize+\height*\cellsize);
        
        \draw[thick, fill=gray] (1*\cellsize, 1*\cellsize) rectangle ++(2*\cellsize,\cellsize);
        \draw[thick, fill=gray] (4*\cellsize, 1*\cellsize) rectangle ++(3*\cellsize,\cellsize);
        \node[label={[centered,below=-6pt,a2]270:$\agentDelayed$}] at (2.5*\cellsize, 0.7*\cellsize) (a2) {\scalebox{\iconscale}{\color{a2}{\faRobot}}};
        \draw[->, a2, thick] (2.6*\cellsize, 0.8*\cellsize) -- (3.2*\cellsize, 0.8*\cellsize) -- (3.2*\cellsize, 2.7*\cellsize) -- node[at end, a2, left=-5pt, label={[below=-7pt, black]270:{$\nodeThree$}}] {$\times$} (2.7*\cellsize, 2.7*\cellsize);
        \node[label={[centered,below=-6pt,a3]270:$\agentNoDelay$}] at (7.5*\cellsize, 0.7*\cellsize) (a3) {\scalebox{\iconscale}{\color{a3}{\faRobot}}};
        \draw[->, a3, thick] (7.2*\cellsize, 0.8*\cellsize) -- (7.2*\cellsize, 2.4*\cellsize) -- node[at end, a3, left=-5pt] {$\times$} (5.5*\cellsize, 2.4*\cellsize);
        \node[label={[centered,below=-6pt,a4]270:$\agentWithFlexibility$}] at (7.5*\cellsize, 1.7*\cellsize) (a4) {\scalebox{\iconscale}{\color{a4}{\faRobot}}};
        \draw[->, a4, thick] (7.6*\cellsize, 1.7*\cellsize) -- (7.6*\cellsize, 2.7*\cellsize) -- (3.8*\cellsize, 2.7*\cellsize) -- (3.8*\cellsize, 0.7*\cellsize) -- node[at end, a4, right=-5pt, label={[below=-7pt, black]270:{$\nodeTwo$}}] {$\times$} (6.5*\cellsize, 0.7*\cellsize); 
        \node at (3.5*\cellsize,2.45*\cellsize) (w) {$\nodeCritical$};
        \node at (4.2*\cellsize,0.5*\cellsize+\height*\cellsize) (u) {$\nodeOne$};
    \end{tikzpicture}
    \caption{Example Multi-Agent Pathfinding warehouse problem. Each corridor has a width of one.}
    \label{fig:ex}
\end{figure}

Figure~\ref{fig:ex} shows a delay-replanning problem example with three agents starting at~$\timeVar=0$ that must arrive before~$\timeVar~=~12$ at their goal.
While this shows a discretized space and time, the model and algorithms are continuous.
Agents~$\agentDelayed$ and~$\agentWithFlexibility$ both use the center corridor, where agent~$\agentDelayed$ is set to go first.
Suppose that agent~$\agentDelayed$ is delayed for departure at~$\timeVar=0$, but we do not know how much.
Agent~$\agentWithFlexibility$ is set to enter the corridor at~$\timeVar=5$ and exit at~$\timeVar=8$, so agent~$\agentDelayed$ can no longer use the corridor before agent~$\agentWithFlexibility$ without delaying it.
Agent~$\agentWithFlexibility$ has two time steps of flexibility (arrive before~$\timeVar=12$) from location~$\nodeOne$ on, because before that~$\agentNoDelay$ is right behind and should not be affected.
Departing by~$\timeVar=2$, agent~$\agentDelayed$ can cross the corridor as planned.
When departing during~$\timeVar~\in~[2, 4\rangle$, agent~$\agentDelayed$ can still go first, so the order remains the same, because agent~$\agentWithFlexibility$ can use its flexibility at~$\nodeOne$ departing there at~$\timeVar=7$.
After~$\timeVar=4$, agent~$\agentDelayed$ must wait for agent~$\agentWithFlexibility$ and then use the middle corridor, departing earliest at~$\timeVar=7$; or agent~$\agentDelayed$ can reroute through the left corridor, which is a longer path, so this is only faster when departing during~$\timeVar \in [4,5\rangle$.
This example shows the optimal routes based on the departure time of agent~$\agentDelayed$, resulting in the tipping point at~$\timeVar=4$: if agent~$\agentDelayed$ leaves before this time, then it can go first (possibly using~$\agentWithFlexibility$'s flexibility); otherwise agent~$\agentWithFlexibility$ should go first (and agent~$\agentDelayed$ may reroute through the left corridor).

This work is closely related to Multi-Agent Execution Delay Replanning (MAEDeR) \cite{Hanou2024}, which addresses a stricter subset of delay replanning problems, where the originally delayed agent must recover without affecting the plans of any other agents.
Like MAEDeR, our approach uses the Safe Interval Path Planning (SIPP) state space to encode when resources are not used by other agents \cite{Phillips2011}.
With this problem representation, optimal single-agent plans for all start times can be efficiently computed \cite{Thomas2023}, by treating other agents as moving obstacles.
\citet{Hanou2024} used this to shift replanning to be `in advance', and recover from some execution conflicts instantly by looking up a new plan for the delayed agent in its any-start-time plan.
While the resulting multi-agent plan is safe, MAEDeR does not necessarily provide an optimal solution, as it assumes that the plans of agents other than the delayed agent cannot be modified.

Compared to MAEDeR and SIPP-based replanning, because we allow delaying other agents, we can reduce overall delay. 
Our method, called FlexSIPP, still provides near-instant delay replanning, and the required precomputation runs in polynomial time. 
FlexSIPP was originally described by \citet{Kemmeren2025} as a railway-specific approach.
In this paper, we formalize FlexSIPP and generalize it to a multi-agent, continuous-time, path-replanning setting, showing that we can efficiently replan a single agent when it is delayed or its initial plan fails for other reasons.
FlexSIPP provides two use cases: precompute the any-start-time plan for any delayed agent to determine the tipping point, or quickly handle a single delay by returning a plan that minimizes total delay.
If only delaying agents with flexibility does not yield a plan, we quickly determine so and require full replanning.

Our main contribution is the FlexSIPP algorithm, which quickly replans a delayed agent in a multi-agent pathfinding context by sequentially handling delays. 
We present a method to determine agents' flexibility and use tipping points to illustrate the replanning trade-off between delaying other agents and further delaying or rerouting a delayed agent.
We evaluate FlexSIPP on the MovingAI MAPF benchmark and a case study of the Dutch railway network, demonstrating the benefits of automatically identifying tipping points and using them in real-world situations.

\section{Background}
\label{sec:back}

We begin by describing the state-space representation of Safe Interval Path Planning (SIPP); Any-Start-Time Planning, which is the single-agent optimal search used by our precomputation; and Multi-Agent Execution Delay Replanning~(MAEDeR), which is the most closely related formulation of the multi-agent delay replanning problem.

\subsection{Safe Interval Path Planning}
SIPP \cite{Phillips2011} is a single-agent state-space search problem defined by a tuple~$\langle \states, \SIPPedges, \edgeDurationVar, \startConfig{}, \problemStartTime, \goalConfig{} \rangle$, 
where each state~$\statedef$ is a configuration~$\configuration$ (e.g., agent location) and a \emph{safe interval}~$\intervaldef$, which is a continuous timespan from~$\timeStart$ to~$\timeEnd$ when it is safe for the agent to be in configuration~$\configuration$. 
An edge~$\edgedef~\in~\SIPPedges$ for states~$\stateOne, \stateTwo \in \states$ denotes the safe interval~$\interval$ when the edge~$(\stateOne,\stateTwo)$ can be used by the agent, and the cost of this edge is defined by~$\edgeDuration{\stateOne,\stateTwo}$, which is the minimum time it takes an agent to traverse the edge. 
A SIPP graph is thus similar to a time-expanded graph, representing all intervals during which edges can be safely traversed. 
The objective of a SIPP problem is to find the path of minimal duration~$\edgeDurationVar$ from the starting configuration~$\startConfig{}$ at the starting time~$\problemStartTime$ to a state with the goal configuration~$\goalConfig{}$.
One configuration may have many SIPP states with different (non-overlapping) safe intervals.
A SIPP state can also have several SIPP edges connected to another SIPP state.
For example, in the scenario in Figure~\ref{fig:ex}, agent~$\agentDelayed$ has one SIPP state in its final configuration~$\nodeThree$ with an infinite interval, but the previous configuration~$\nodeCritical$ on its path originally has two SIPP states because this configuration is only safe for~$\agentDelayed$ to be in before or after agent~$\agentWithFlexibility$ crosses it.
Both these states connect to the SIPP state on configuration~$\nodeThree$ with a different SIPP edge, allowing agent~$\agentDelayed$ to move from~$\nodeCritical$ to~$\nodeThree$ before or after agent~$\agentWithFlexibility$.

\begin{figure}[t]
    \centering
    \begin{tikzpicture}[scale=0.25]
    \draw[thick,->] (0,0) -- (6.5,0) node[anchor=north west] {$t_{depart}$};
    \draw[thick,->] (0,2.5) -- (0,3.5);
    \draw[thick] (0,0) -- (0,1.5);
    \draw[thick,dashed] (0,1.5) -- (0,3);
    \draw (1.5 cm,5pt) -- (1.5 cm,-5pt) node[anchor=north] {$\zeta$};
    \draw (3 cm,5pt) -- (3 cm,-5pt) node[anchor=north] {$\alpha$};
    \draw (4.5 cm,5pt) -- (4.5 cm,-5pt) node[anchor=north] {$\beta$};
    \draw (1pt,3 cm) -- (-1pt,3 cm) node[anchor=east] {$\infty$};
    \draw (5pt, 1) -- (-5pt, 1) node[anchor=east] {$\alpha + \Delta$};
    \draw (-1.5, 3) -- (-1.5, 3) node[anchor=east] {$t_{arrive}$};
    \filldraw[black] (1.5,1) circle (2pt);
    \filldraw[black] (3,1) circle (2pt);
    \filldraw[black] (4.5,2.2) circle (2pt);
    \draw (0,3) -- (1.5,3);
    \draw [dashed] (1.5,3) -- (1.5,1);
    \draw (1.5,1) -- (3,1);
    \draw (3,1) -- (4.5,2.2);
    \draw [dashed] (4.5,2.2) -- (4.5,3);
    \draw (4.5,3) -- (6,3);
    \end{tikzpicture}
    \caption{An ATF with parameters $\zeta,\alpha,\beta$, and $\Delta$.}
    \label{fig:atf}
\end{figure}

\subsection{Any-Start-Time Planning}
The original SIPP algorithm is an~$A^*$ search with the scalar earliest arrival time at a state as the cost ($g$-value) of the path to that location.
Any-start-time planning (@SIPP) generalizes SIPP to plan for all start times by searching with earliest arrival-time functions (ATFs), rather than scalar $g$-values \cite{Thomas2023}.
This creates the benefit that we keep track of all possible plans, including different start times, arrival times, and paths.
An ATF maps the departure time of the agent, when it begins traversing an edge, to the earliest arrival time at the end of the edge. 
A SIPP graph~$\langle \states, \SIPPedges, \edgeDurationVar \rangle$ is transformed to an @SIPP graph~$\langle \states, \SIPPedges, \atfFunc[\edge] \rangle$ by generating the arrival time functions~$\atfFunc[\edge]$ for all edges~$\edge \in \SIPPedges$.
For each~$\edgedef$, the safe intervals from the source state~$\stateOne$ (interval~$[ \timeStart^{\stateOne}, \timeEnd^{\stateOne} \rangle$), the destination state~$\stateTwo$ (interval~$[\timeStart^{\stateTwo}, \timeEnd^{\stateTwo} \rangle$), and the edge interval~$\interval = [\timeStart^{\edge},\timeEnd^{\edge} \rangle$ are compiled into one ATF.
This ATF~$\atfFunc[\edge]$ is defined from the following four parameters:
\begin{align}
    &\earliestWaitTime = \timeStart^{\stateOne} \label{eqn:zetacompute}\\
    &\begin{aligned}
    \earliestStartTime = \max(&\timeStart^{\edge}, \timeStart^{\stateOne}, \timeStart^{\stateTwo} - \edgeDuration{\stateOne, \stateTwo}) \label{eqn:alphacompute} 
    \end{aligned}\\
    &\begin{aligned}
    \unsafeTime = \min(&\timeEnd^{\edge}, \timeEnd^{\stateOne},  \timeEnd^{\stateTwo} - \edgeDuration{\stateOne, \stateTwo}) \label{eqn:betacompute}
    \end{aligned}\\
    &\transitTime = \edgeDuration{\stateOne, \stateTwo},\label{eqn:deltacompute}
\end{align}
where~$\earliestWaitTime$ is the earliest time the agent can safely wait at the starting state~$\stateOne$,~$\earliestStartTime$ is the earliest time it is safe for the agent to depart state~$\stateOne$,~$\unsafeTime$ is the time at which the edge becomes unsafe to start traversing, and~$\transitTime$ is the time it takes the agent to traverse the edge \cite{Thomas2023}.
The ATF~$\atfFunc[\edge]$ is then a piecewise linear function constructed by~$\earliestWaitTime,\earliestStartTime,\unsafeTime,\transitTime$:
\begin{equation}\label{eq:atf}
    \atfFunc[\edge](\timeVar) = 
    \begin{cases}
        \infty & \timeVar < \earliestWaitTime \\
        \earliestStartTime + \transitTime & \earliestWaitTime \leq \timeVar < \min(\earliestStartTime, \unsafeTime)\\
        \timeVar + \transitTime & \earliestStartTime \leq \timeVar < \unsafeTime \\
        \infty & \unsafeTime \leq \timeVar.
    \end{cases}
\end{equation}
Figure~\ref{fig:atf} visualizes such an arrival time function. 
Equation~\ref{eq:atf} considers the ATF of an edge~$\atfFunc[\edge]$, which can be combined into an ATF for an entire path~$\atfFunc[\pathParam]$ \cite{Thomas2023}.
Moreover, the~$\min$ in the second case is because~$\unsafeTime < \earliestStartTime$ is possible whenever the plan of a path~$\pathParam$ requires more waiting than feasible before the initial state becomes unsafe.
The duration~$\transitTime$ in the edge ATF~$\atfFunc[\edge]$ is simply the duration~$\edgeDurationVar$ of~$\edge$; however, for a path ATF~$\atfFunc[\pathParam]$,~$\transitTime$ is the entire path's duration.

\subsection{Multi-Agent Execution Delay Replanning}
We build on the work of \citet{Hanou2024}, which applied Any-Start-Time Planning to solve a subset of multi-agent delay replanning problems. 
They defined the Multi-Agent Execution Delay Replanning (MAEDeR) problem, which is a single-agent problem in the multi-agent setting.
A MAEDeR problem is defined by the tuple~$\maeder$. 
Here,~$\network$ is the set of components that can be represented as a graph with edges between locations (e.g., a railway network),~$\agents$ is the set of agents (e.g., trains) navigating through the network, and~$\chars$ holds the problem characteristics on how agents interact with the network and each other (e.g., speed).
A solution to MAEDeR is a function~$\maedersol$ that takes an agent~$\delayFunc{\agent} \in \agents$ and a positively delayed start time~$\delayFunc{\timeVar}$, and returns a shortest safe plan for this delayed agent, without affecting the plans of other agents in~$\agents$; or returns a failure if no such plan is feasible and fuller replanning is required. 

\citet{Hanou2024} described two algorithms to solve MAEDeR problems. 
First, Replanning SIPP performs a SIPP search for the delayed agent, yielding a set of SIPP graphs that can be generated before execution but are searched during execution. 
Second, @MAEDeR precomputes the SIPP graphs and transforms them into @SIPP graphs (with ATF edges).
These are searched before execution using the RePEAT algorithm \cite{Thomas2023}, which repeatedly searches for optimal plans for monotonically increasing start times.
The resulting set of plans can be rapidly queried for an optimal plan corresponding to any departure time.
Essentially, @MAEDeR replans in advance, so that at execution time, replanning just involves querying the any-start-time plan of the delayed agent.
While they find the optimal solution provided only the delayed agent is replanned, the resulting multi-agent plan is not necessarily optimal.
Sometimes, a lower-cost overall plan can be found by also replanning other agents.

\paragraph{Comparison}
The SIPP graph inherently encodes the safe intervals, allowing the delayed agent to choose a time in this interval during which this agent can move between configurations without changing anything regarding the other agents. 
Any-start-time planning has the additional benefit of precomputing routes for all possible delays for this single agent. 
@MAEDeR precomputes these any-start-time plans to instantly recover from delays, which can be repeated for multiple delays due to quick updating. 
However, using the possibility of delaying other agents' plans to provide more or better solutions remains a gap in this line of research.

\section{\problem{}}
\label{sec:def}
Consider a multi-agent path planning problem and a solution to this problem consisting of trajectories for all agents, and assume that the plan of one agent cannot be safely executed (e.g., the agent is delayed); we want to replan that agent, using the temporal flexibility of the other agents.
We start with the model of this replanning problem as an~@SIPP graph~\cite{Thomas2023}.
This graph is created from the perspective of the delayed agent, which means that the safe intervals are safe for the delayed agent to not conflict with the movements of all the other agents.
Later, we explain how to extend this graph with the temporal flexibility of the other agents.
We use the arrival time functions as costs to keep track of different paths to the goal, where some paths may use flexibility and others may not.
Formally, a \emph{\problem{} problem} is defined as the tuple~$\langle \agents,  \delayFunc{\agent}, \network, \edgeDurationVar, \maxTime, \sol \rangle$, where
\begin{itemize}
    \item $\agents$ defines a set of~$\numAgents$~agents,
    \item $\delayFunc{\agent} \in \agents$ is the \emph{delayed agent}, orignally starting at~$\startTime{}$,
    \item $\network$ is the set of components that can be represented as an @SIPP graph~$\langle \states, \SIPPedges, \atfFunc[\edge] \rangle$ for agent~$\delayFunc{\agent}$ where all agents~$\agent \neq \delayFunc{\agent} \in \agents$ are considered dynamic obstacles,
    \item $\edgeDuration{\configuration_1,\configuration_2}$ gives the duration of an edge: the minimum travel time from configuration~$\configuration_1$ to configuration~$\configuration_2$,
    \item $\maxTime$ gives the global end time of the scenario,
    \item $\sol$ are safe and valid trajectories; a trajectory~$\plan{\agent} \in \sol$ is a sequence of~$\planLength{\agent}$ configuration/time pairs~$(\configuration_\stateIndex, \timeVar_\stateIndex)$ specifying that agent~$\agent$ arrives at configuration~$\configuration_\stateIndex$ at time~$\timeVar_\stateIndex$,
    with start~$\configuration_1 = \startConfig{\agent}$ and goal~$\configuration_{\planLength{\agent}} = \goalConfig{\agent}$, 
    for each agent~$\agent \in \agents$.
\end{itemize}
A trajectory $\plan{\agent}$ of length~$\planLength{\agent}$ is valid if and only if the network~$\network$ contains every connection between subsequent configurations, and the move respects the minimum travel time: $\forall \stateIndex < \planLength{\agent}:(\configuration_\stateIndex,\configuration_{\stateIndex+1}) \in \network~\text{and}~\timeVar_\stateIndex + \edgeDuration{\configuration_\stateIndex,\configuration_{\stateIndex+1}} \leq \timeVar_{\stateIndex+1}$.
The set of trajectories~$\sol$ is safe if and only if no two agents occupy the same configuration or edge simultaneously.
Where a \emph{path} concerns only the physical locations, a \emph{trajectory} is a time-stamped path, the states for an agent's \emph{plan}.
A delayed agent may be \emph{rerouted}, changing the path, while other agents can only be \emph{delayed}, keeping the same path but changing the timing.
The \problem{} problem has two use cases with different objectives. 
The first is to identify the tipping points. 
The second is to retrieve the optimal solution for a given start time of the delayed agent, where other agents may use their temporal flexibility.
While correlated with minimizing makespan, this is not our objective.
We define an agent's temporal flexibility as:

\begin{definition}[Flexibility]
\label{def:flex}
    The \emph{flexibility}~$\flex{\agent}{\configuration_\stateIndex, \timeVar_\stateIndex}$ of an agent~$\agent \neq \delayFunc{\agent} \in \agents$ in configuration~$\configuration_\stateIndex$ where it arrives at time $\timeVar_\stateIndex$ is the amount of time it can \emph{safely} be delayed in that configuration, while ensuring that 1) all other agents keep their original trajectories, and 2) all agents visit the configurations in the order defined by these original trajectories.
\end{definition}

Algorithm~\ref{alg:flex} gives a polynomial-time algorithm to compute~$\flex{\agent}{\configuration_\stateIndex,\timeVar_\stateIndex}$.
This is, in fact, equivalent to (backward) propagating the temporal constraints on agent~$\agent$’s configuration~$\configuration_\stateIndex$ at a time~$\timeVar_\stateIndex$ as common in Simple Temporal Networks (STNs) \cite{dechterTemporalConstraintNetworks1991}.
For each configuration visited by both agent~$\agent$ and some other agent~$\agentOther$ after~$\timeVar_\stateIndex$~(line 6), we constrain the flexibility to maintain the current visiting order~(line 7-8).
An agent's flexibility is determined by the local flexibility~(line 3) along its future path, based on fixed visiting times of other agents~(line~4).
We compute this for all agents except for the delayed agent.
When precomputing the any-start-time plan for an agent, that agent is regarded as the delayed agent.
Proposition~\ref{thm:flex} states that the temporal flexibility can always be computed efficiently.
Algorithm~\ref{alg:flex} runs in~$O(\numAgents \cdot \planLengthMax^2)$ for each agent, where~$\planLengthMax$ is the maximum length across all trajectories and~$\numAgents$ is the number of agents, thus it is polynomial time, proving Proposition~\ref{thm:flex}. 
The size of the any-start-time plan is linear in the number of agents (each agent can create a new interval), but does not scale with the planning horizon, as intervals are continuous, though runtime does increase linearly.

\begin{proposition}[Flexibility]
\label{thm:flex}
    Given a \problem{} problem, we can derive the flexibility of all agents in polynomial time.
\end{proposition}

\begin{algorithm}[t]
    \begin{algorithmic}[1]
        \State $\flex{\agent}{\goalConfig{\agent}, \timeVar_{\planLength{\agent}}} \gets \maxTime - \timeVar_{\planLength{\agent}}$
        \For{$\stateIndex \in [\planLength{\agent}-1,\dots,1]$} \Comment{Configurations visited by plan~$\plan{\agent}$ in reverse order}
            \State $\waitTimeSteps{}{} \gets \timeVar_{\stateIndex+1} - \timeVar_\stateIndex - \edgeDuration{\configuration_\stateIndex,\configuration_{\stateIndex+1}}$
            \Comment{Local flexibility}
            \State $\tmpFlex \gets \waitTimeSteps{}{} + \flex{\agent}{\configuration_{\stateIndex+1}, \timeVar_{\stateIndex+1}}$
            \For {$\agentOther \in \agents \;|\; \agent \neq \agentOther$}
            \If{agent~$\agentOther$ visits configuration~$\configuration_\stateIndex$ after $\timeVar_\stateIndex$}
                    \State $\timeVar^{\agentOther} \gets$ earliest time that~$\agentOther$ visits~$\configuration_\stateIndex$ after~$\timeVar_\stateIndex$
                    \State $\tmpFlex \gets \min(\tmpFlex,  \timeVar^{\agentOther} - \timeVar_\stateIndex)$
                \EndIf
            \EndFor
            \State $\flex{\agent}{\configuration_\stateIndex, \timeVar_\stateIndex} \gets \tmpFlex$
        \EndFor
    \end{algorithmic}
    \caption{Flexibility for one agent~$\agent \in \agents$ based on the original plan~$\sol$, where~$\planLength{\agent}$ is the length of its plan~$\plan{\agent}$.}
    \label{alg:flex}
\end{algorithm}

\begin{example}[Flexibility]
\label{ex:prob}
    Consider replanning agent~$\agentDelayed$ in Figure~\ref{fig:ex} again.
    We derive the flexibility of agent~$\agentWithFlexibility$ along its path to know if we can slightly delay~$\agentWithFlexibility$ to let agent~$\agentDelayed$ use the middle corridor first.
    The first part of the path of~$\agentWithFlexibility$ is also used by agent~$\agentNoDelay$, and the flexibility of~$\agentWithFlexibility$ cannot influence the path of~$\agentNoDelay$, so agent~$\agentWithFlexibility$ has no flexibility here.
    However, beyond~$\agentNoDelay$'s goal, before reaching the corridor, agent~$\agentWithFlexibility$ can wait safely.
    Configuration~$\nodeOne$ is marked along this section, which agent~$\agentWithFlexibility$ originally occupies during~$\timeVar\in[4,5\rangle$.
    We take the scenario to have~$\maxTime=12$, so agent~$\agentWithFlexibility$ originally waits at its goal~$\nodeTwo$ during~$\timeVar\in[10,12\rangle$, yielding a local flexibility of~$\flex{\agentWithFlexibility}{\nodeTwo,10}=2$.
    For each configuration visited by agent~$\agentWithFlexibility$ before its goal, the local flexibility is propagated until the point on~$\agentWithFlexibility$'s path also used by~$\agentNoDelay$. 
    The flexibility in the safe point to wait is thus~$\flex{\agentWithFlexibility}{\nodeOne,4}=2$.
\end{example}
We use the flexibility of all agents to identify a possible reordering between a delayed agent~$\delayFunc{\agent}$ and another agent~$\agentOther$ visiting a configuration. The smallest delay at which a reordering improves the objective is called a tipping point:

\begin{definition}[Tipping Point]
\label{def:tip}
    Take agents~$\agent$ and~$\agentOther$ and their original plans~$\plan{\agent},\plan{\agentOther}$, where agent~$\agent$ uses some configuration~$\tippingLoc$ before agent~$\agentOther$ uses this~$\tippingLoc$.
    The~\emph{tipping point} is the pair~$\tipping$, such that until this time~$\tippingTime$ (not included) the ordering of the agents using this configuration~$\tippingLoc$ is agent~$\agent$ before agent~$\agentOther$, if necessary by delaying~$\agentOther$. 
    After this time~$\tippingTime$, the order of the agents is swapped, so $\agentOther$ uses configuration~$\tippingLoc$ first (at its scheduled time) and agent~$\agent$ waits until the configuration becomes available again.
\end{definition}

\begin{example}[Tipping Point]
\label{ex:tip}
    Agent~$\agentDelayed$ needs to replan and can use some of agent~$\agentWithFlexibility$'s flexibility to still traverse first.
    We can delay agent~$\agentWithFlexibility$ to wait in configuration~$\nodeOne$ during~$\timeVar \in [4,7\rangle$ where~$\timeVar\in[4,5\rangle$ is its scheduled time, and in the rest of the interval it uses some flexibility~$\flex{\agentWithFlexibility}{\nodeOne,4}=2$.
    If agent~$\agentDelayed$ uses the middle corridor before agent~$\agentWithFlexibility$, then it must clear this before~$\timeVar=8$, or equivalently, it has to depart its start configuration at the latest by~$\timeVar=4$, reaching configuration~$\nodeCritical$ at~$\timeVar=7$.
    If agent~$\agentDelayed$ reaches~$\nodeCritical$ \emph{before}~$\timeVar=7$, then agent~$\agentDelayed$ can cross the corridor first; otherwise, agent~$\agentWithFlexibility$ should move first at its scheduled time. 
    We thus have a tipping point~$\tipping=(\nodeCritical,7)$ for agents~$\agentDelayed$ and~$\agentWithFlexibility$.
\end{example}

\section{FlexSIPP}
\label{sec:meth}
This section presents the FlexSIPP algorithm, which solves the \problem{} problem and identifies the tipping points.
For any configuration-time pair~$(\configuration_\stateIndex, \timeVar_\stateIndex)$ in an agent's plan~$\plan{\agent}$, Algorithm~\ref{alg:flex} gives the flexibility in that configuration, which is a scalar~$\flex{\agent}{\configuration_\stateIndex, \timeVar_\stateIndex}$. 
We represent this flexibility in the search by creating an additional safe edge on any edge~$(\configuration',\configuration_\stateIndex)$ in our @SIPP graph for agent~$\delayFunc{\agent}$, so that it can be used for an alternative plan for this delayed agent. 
This tracks if the flexibility of an agent is used, by updating the configuration-time pairs for the plan of the agent~$\agent$ whose flexibility we have used.
Just as with the other @SIPP edges, this edge has an arrival-time function.
We construct the \emph{flexible ATF}~$\flexFunc{\atfFunc}$ based on an original ATF~$\atfFunc$ of agent~$\delayFunc{\agent}$ by changing only the latest start time~$\unsafeTime$ to~$\flexFunc{\unsafeTime}$ to add the flexibility.
This ATF~$\flexFunc{\atfFunc}$ uses the flexibility of the first agent~$\flexFunc{\agent}$ that visits the respective configuration after the delayed agent~$\delayFunc{\agent}$ does, unless the flexibility~$\flex{\flexFunc{\agent}}{\configuration,\timeVar}$ is zero.
For every edge~$\edge$ in the @SIPP graph with ATF~$\atfFunc[\edge]$ for agent~$\delayFunc{\agent}$, we create the~$\flexFunc{\atfFunc}[\edge,\flex{\flexFunc{\agent}}{\configuration,\timeVar}]$ with~$\flexFunc{\unsafeTime} = \unsafeTime + \flex{\flexFunc{\agent}}{\configuration,\timeVar}$ based on the original ATF: 
\begin{equation}\label{eq:atff}
    \flexFunc{\atfFunc}[\edge, \flex{\flexFunc{\agent}}{\configuration,\timeVar}](\timeVar) = 
    \begin{cases}
        \infty & \timeVar < \earliestWaitTime \\
        \earliestStartTime + \transitTime & \earliestWaitTime \leq \timeVar < \min(\earliestStartTime, \flexFunc{\unsafeTime})\\
        \timeVar + \transitTime & \earliestStartTime \leq \timeVar < \flexFunc{\unsafeTime} \\
        \infty & \flexFunc{\unsafeTime} \leq \timeVar.
    \end{cases}
\end{equation}

\begin{example}[ATF]
\label{ex:atf}
    Agent~$\agentDelayed~(\delayFunc{\agent})$ is replanning a path through the middle corridor, maybe delaying agent~$\agentWithFlexibility~(\flexFunc{\agent})$.
    We consider the~@SIPP edge~$\edge_{\nodeCritical}$ representing the agent~$\agentDelayed$'s move from the corridor to configuration~$\nodeCritical$.
    For this edge's ATF, we do not worry about other conflicts along the path; this is handled by searching for a complete path between safe configurations.
    Agent~$\agentWithFlexibility$ originally occupies configuration~$\nodeCritical$ during~$\timeVar \in [5,6\rangle$.
    Agent~$\agentDelayed$ can either take edge~$\edge_{\nodeCritical}$ before~$\agentWithFlexibility$, represented by~$\atfFunc_1[\edge_{\nodeCritical}]$ constructed by~$\earliestWaitTime=\earliestStartTime=0$, $\unsafeTime=3$ (the latest~$\agentDelayed$ has to start moving to~$\nodeCritical$ to arrive before~$\agentWithFlexibility$), and~$\transitTime=1$ (just one cell), or after~$\agentWithFlexibility$, which is represented by ATF~$\atfFunc_2[\edge_{\nodeCritical}]$ constructed by~$\earliestWaitTime=7$ (the earliest it can be here after~$\agentWithFlexibility$),~$\earliestStartTime=7$,~$\unsafeTime=\maxTime-1=11$, and~$\transitTime=1$.
    However, we have another option (besides rerouting through the left corridor) to let agent~$\agentWithFlexibility$ wait, using its flexibility~$\flex{\agentWithFlexibility}{\nodeOne,4}=2$ (Example~\ref{ex:prob}).
    In this case, agent~$\agentWithFlexibility$ uses~$\nodeCritical$ during~$\timeVar\in[7,8\rangle$, and agent~$\agentDelayed$ thus has a flexible ATF~$\flexFunc{\atfFunc}[\edge_{\nodeCritical}]$ constructed by~$\earliestWaitTime=0$,~$\earliestStartTime=3$, and~$\flexFunc{\unsafeTime}=3+2=5$ (updating~$\unsafeTime$ from~$\atfFunc_1[\edge_{\nodeCritical}]$), and~$\transitTime=1$.
\end{example}

\subsection{Search}
We find the any-start-time plan for agent~$\delayFunc{\agent}$ using the RePEAT algorithm \cite{Thomas2023} on the @SIPP graph with the additional flexible ATFs.
These ATFs define the flexibility that can be used during the search.
The only update in the search procedure is the tracking of flexibility.
When the delayed agent~$\delayFunc{\agent}$ uses a flexible ATF~$\flexFunc{\atfFunc}$ to construct its new path, this means that another agent~$\flexFunc{\agent} \neq \delayFunc{\agent}$ has to use some of its flexibility, which thus impacts that agent's plan.
To ensure feasible plans for the other agents, this impact must be accounted for.
When the agent~$\flexFunc{\agent}$ is delayed in some configuration~$\configuration$ by at most~$\flex{\flexFunc{\agent}}{\configuration,\timeVar}$, every configuration-time pair in the single-agent plan~$\plan{\flexFunc{\agent}}$ that is visited after configuration~$\configuration$ must now be updated. 
Each such configuration~$\overline{\configuration}$ was originally safe during some interval~$[ \underline{\timeStart},\underline{\timeEnd}\rangle$ (before agent~$\flexFunc{\agent}$ visits), and also safe during some interval~$[ \overline{\timeStart},\overline{\timeEnd} \rangle$ (after agent~$\flexFunc{\agent}$ visits). 
These safe intervals are then updated to~$[ \underline{\timeStart},\underline{\timeEnd}+\flex{\flexFunc{\agent}}{\configuration,\timeVar}\rangle$ and~$[ \overline{\timeStart}~+~\flex{\flexFunc{\agent}}{\configuration,\timeVar},~\overline{\timeEnd}\rangle$, respectively.

\begin{example}[Search with flexibility]\label{ex:search}
    We show how the arrival time functions change during search when using flexibility.
    In this case, agent~$\flexFunc{\agent}=\agentWithFlexibility$ and~$\delayFunc{\agent}=\agentDelayed$.
    We have flexibility~$\flex{\agentWithFlexibility}{\nodeOne,4}=2$ and take~$\maxTime=12$.
    The path ATF~$\flexFunc{\atfFunc_1}[\pathParam]$ is constructed by~$\earliestWaitTime=0, \earliestStartTime=2$,~$\flexFunc{\unsafeTime}=4$ and~$\transitTime=4$, because it has to clear the corridor before agent~$\agentWithFlexibility$ uses it (before~$\timeVar=2$ no flexibility is used).
    Agent~$\agentWithFlexibility$ has a path ATF~$\atfFunc_2[\pathParam]$ constructed by~$\earliestWaitTime=0$,~$\earliestStartTime=2$,~$\unsafeTime=1$, and~$\transitTime=10$, which means it has to depart its start configuration before~$\unsafeTime=1$ (because agent~$\agentNoDelay$ is right behind), but has flexibility along its path to wait at most~$\earliestStartTime=2$, while the plan without waiting takes~$\transitTime=10$.
    Assume agent~$\agentDelayed$ uses the corridor first, starting at~$\timeVar=3$, forcing agent~$\agentWithFlexibility$ to wait until~$\timeVar=7$ to enter the corridor from configuration~$\nodeOne$, taking a delay of 2 by using some of its flexibility (delay can be~$<3$ but we show 2 because of discretization).
    If we consider the safe intervals on configuration~$\nodeCritical$, which were originally safe during~$[ \underline{0},\underline{4}\rangle$ (before~$\agentWithFlexibility$ visits as planned) and~$[ \overline{6},\overline{\maxTime}\rangle$ (after~$\agentWithFlexibility$ visits as planned), these are updated to~$[ \underline{0},\underline{6}\rangle$ and~$[ \overline{7},\overline{\maxTime}\rangle$, accounting only for~$\agentWithFlexibility$'s moves as we are replanning~$\agentDelayed$ and agent~$\agentNoDelay$ does not use the corridor.
\end{example}

\begin{figure}[t]
    \centering
    \begin{tikzpicture}
        \begin{axis}[
            axis lines=left,
            ylabel={Arrival Time~$\agentDelayed$},
            xlabel={Departure Time~$\agentDelayed$},
            xtick={0,2,4,6,8},
            ytick={0,2,4,6,8,10,12},
            xmin=-0, xmax=8,
            ymin=0, ymax=12,
            tick style={draw=black},
            grid=major,
            grid style={dashed, gray!30},
            width=\columnwidth,
            height=3.6cm,
            legend style={at={(0.6,0.04)},anchor=south west},
            ylabel style={yshift=-0.2cm},
            xlabel style={yshift=0.2cm},
        ]
        \addplot[col2, very thick] coordinates {(0,4) (2,6)};
        \addlegendentry{FlexSIPP};
        \addplot[col1, loosely dashed, very thick] coordinates {(0,4) (2,6)};
        \addlegendentry{@MAEDeR};
        \node[col2] at (0.6,6) {$\atfFunc_1[\pathParam]$};
        \addplot[col2, loosely dashed, very thick] coordinates {(1,5) (4,8)};
        \node[col2, right] at (4,8) {$\flexFunc{\atfFunc}_1[\pathParam]$};
        \addplot[col2, very thick] coordinates {(4,10) (5,11)};
        \node[col2, left] at (4.6,11) {Reroute};
        \addplot[col2, very thick] coordinates {(5,11) (7,11) (8,12)};
        \node[col2, below] at (6,11) {Wait for~$\agentWithFlexibility$};
        \addplot[col1, loosely dashed, very thick] coordinates {(2,8) (4,10) (5,11) (7,11) (8,12)};
      \end{axis}
    \end{tikzpicture}
    \caption{Results on warehouse example from Figure~\ref{fig:ex}.}
    \label{fig:atf-exp}
\end{figure}

\begin{example}[Any-start-time plan]
    Figure~\ref{fig:atf-exp} shows the results from FlexSIPP for the running example.
    We see that FlexSIPP finds four paths for agent~$\agentDelayed$:~1)~$\atfFunc_1[\pathParam]$ lets agent~$\agentDelayed$ depart until~$\timeVar=1$ without further delays;~2)~$\flexFunc{\atfFunc_1}[\pathParam]$ forces agent~$\agentWithFlexibility$ to use its flexibility;~3)~the path reroutes through the left corrirdor, but after~$\timeVar~=~5$ it is equally fast to wait for agent~$\agentWithFlexibility$, which is preferred as the path stays the same; and~4)~$\agentDelayed$ can only depart after~$\agentWithFlexibility$ has cleared the corridor (during~$\timeVar\in[5,7\rangle$ arrival time does not change), and must depart before~$\timeVar~=~8$ to its goal in time~($\maxTime=12$). 
    This shows the tipping point at~$\timeVar=4$ for starting time (see Example~\ref{ex:tip}).
    The paths found by @MAEDeR only differ during~$\timeVar \in [2,4\rangle$ because flexibility cannot be used.
\end{example}

Using the search procedure described above, we now show that we find the best replanning solution given the agent's flexibility in polynomial time (Proposition~\ref{prop:sol}). 
The flexibility is derived in polynomial time (Proposition~\ref{thm:flex}), and the construction of the @SIPP graph is also done in polynomial time. 
The original @SIPP graph~$\langle \states, \SIPPedges, \atfFunc[\edge] \rangle$ based on the network~$\network$ and set of agent~$\agents$ is of order~$O(|\network| \cdot|\agents|)$ \cite{Hanou2024}.
FlexSIPP adds more safe intervals, but at most one per edge in the original @SIPP graph, so the @SIPP graph is still in the order~$O(|\network| \cdot |\agents|)$.
Each path is split into simple paths with one origin and destination, so we can have several paths for one agent, and the problem size thus scales linearly with the number of simple paths.
The branching factor in the~$A^*$ search is bound by the number of agents; so, it runs polynomially in the number of agents. 
\begin{proposition}
\label{prop:sol}
    If a replanning solution exists that involves only delaying one of the other agents according to their flexibility (Definition~\ref{def:flex}) and possibly rerouting the delayed agent, then we find this solution in polynomial time.
\end{proposition}

\paragraph{Tipping Points}
FlexSIPP returns an any-start-time plan for agent~$\delayFunc{\agent}$ and the tipping points with other agents, meeting the first objective of~\problem{}.
During search, we track the flexibility used of any agent~$\flexFunc{\agent} \neq \delayFunc{\agent}$. 
For each plan found in the any-start-time plan, with a path ATF~$\flexFunc{\atfFunc}[\pathParam]$ using flexibility, the tipping point is then the~$\flexFunc{\unsafeTime}$ parameter, which is the latest time that the agent~$\delayFunc{\agent}$ can still safely use the configuration, based on the flexibility of~$\flexFunc{\agent}$.
For the entire path of~$\agentDelayed$, we have~$\flexFunc{\unsafeTime}=4$ (Example~\ref{ex:search}), which ensures that both agents~$\agentWithFlexibility$ and~$\agentDelayed$ still arrive before~$\maxTime=12$.
So, we see indeed the tipping point identified in Example~\ref{ex:tip}, placed on the starting time of agent~$\agentDelayed$.
The location of the tipping point is the first point along the delayed agent's path where the other agent passes, which is configuration~$\nodeCritical$.
To find the tipping point time stamp, we take the~$\unsafeTime+\transitTime$ of that ATF.
For the ATF~$\atfFunc_1[\edge_{\nodeCritical}]$ of~$\agentDelayed$ moving to~$\nodeCritical$, we had~$\flexFunc{\unsafeTime}=6$ and~$\transitTime=1$ (Example~\ref{ex:atf}), resulting also in the tipping point~$\tipping=(\nodeCritical,7)$. 

\paragraph{Minimize Total Delay}
FlexSIPP can also be used in the second use case of \problem{}: retrieving the optimal solution for a given start time of the delayed agent.
We compute the total delay by comparing each agent's arrival time at its goal with its originally scheduled arrival time.
Given the any-start-time plan, we can derive both the fastest plan for the delayed agent and the plan with the least total delay over all agents.

\paragraph{Rescheduling}
For each plan in the resulting any-start-time plan, FlexSIPP determines the agent(s) whose flexibility is used (if any at all), and if so, when and where the waiting time of the other agent(s) must be adjusted.
These agents can only be forced to wait, not to change their path, and this never causes conflicts, because only feasible delays (i.e., not influencing another agent) are used as input to the search.
The any-start-time plan with flexibility is precomputed before execution, so plan updates occur in constant time and are, in fact, almost instantaneous.
After the update, FlexSIPP can be used to recompute any-start-time plans for any agent, or to simply query the first plan if a new delay is encountered before the any-start-time plan is recomputed.
The safe intervals are continuous, so upon a delay, the exact time point of the delay can be queried for the correct safe interval to start the new plan from.
FlexSIPP searches over the~@SIPP graph, where each edge is thus an ATF representing a continuous time span for safe traversal between configurations. 
The tipping points are single time points, which are retrieved from the end of the safe traversal~($\unsafeTime$ parameter of the ATF).

\section{Case Study: Rotterdam to Schiphol}
\label{sec:case}
To evaluate the effect of using the flexibility of other agents, we consider the example of rerouting trains in a busy railway network. 
When disturbances occur in railway network operations, trains are delayed, and in some cases, they must be rerouted. 
If the tracks are already occupied close to full capacity, it is very difficult to fit this rerouting into the schedule. 
It thus becomes necessary to adjust the schedule slightly to accommodate delayed trains.
Specifically, we consider the tight schedule in the Netherlands, where train services are frequent and expected to become even more metro-like in the next few years \cite{ProRail2024}[p.21].
The railway network in the Netherlands is mostly double-track, allowing trains to be rerouted to an alternative track.
Then, replanning must consider the trains originally scheduled to use that track, preferably without disrupting their service too much. 

A high-speed line (HSL) runs between Rotterdam Central and Schiphol Airport to facilitate easy travel between two of the largest cities: Amsterdam and Rotterdam.
However, the HSL is particularly prone to issues: train punctuality is only 69\% on this line, compared to 89\% on the rest of the network \cite{NS2024annual}. 
When the HSL is unavailable due to large disruptions, trains scheduled to use it must be rerouted over the regular network, through Leiden and The Hague (Figure~1 in the Appendix). 
In this case, the previously discussed algorithm (@MAEDeR) for the Multi-Agent Execution Delay Replanning problem falls short because the timetable is so tight that no complete path fits between the regular trains.
So, we must slightly delay some of the regular trains to allow for the HSL-rescheduled trains to be scheduled at all.

In practice, railway companies (in the Netherlands) use \emph{train handling documents}, which can be understood as a flow chart saying what the appropriate action is if a train traveling a specific path is delayed \cite{ProRailTAD}. 
These documents thus specify the tipping points, when to delay the other trains, and when the delayed train should allow other trains to proceed first.
For example, in the case of the Rotterdam-Leiden-Schiphol line, the tipping point used in industry for regular intercity and sprinter trains is~7~minutes; regular trains on the path between The Hague and Leiden can thus be delayed by~7~minutes to allow another train to proceed first \cite{Kemmeren2025}[p.14].
Currently, these documents are created manually, although they are not specified for every situation (e.g., this Eurostar case). 
So, FlexSIPP provides a safe and automated alternative.

\section{Evaluation}
\label{sec:eval}
We evaluate our method on the case study and compare it to @MAEDeR, which does not allow delaying other agents.
\begin{enumerate}[label=\textbf{Q\arabic*}]
    \item\label{q:tip} Can FlexSIPP provide tipping points for replanning a high-speed train in the Netherlands?
    \item\label{q:info} Does the any-start-time plan found by FlexSIPP hold more information than the one returned by @MAEDeR
    \item\label{q:time} How does FlexSIPP's runtime compare to @MAEDeR?
    \item\label{q:seq} Can FlexSIPP plan with sequential delays?
\end{enumerate}

\subsection{Experimental Setup}
FlexSIPP is implemented in two stages: generation of the @SIPP graph in Python (implemented for both railways and MAPF instances), and search in C++ that takes an @SIPP graph and computes the any-start-time plan for the delayed agent with the specified amount of flexibility.
FlexSIPP is released as a Python package \cite{FlexSIPPcode2026}.
All experiments were run using the \citet{DHPC2024}; each MAPF experiment had a 3GB memory limit and was run in sequence without a time limit, and the EuroStar experiments were given 32000MB of memory.

\subsection{Case Study: Eurostar Rerouting}
We reroute the Eurostar train on the Rotterdam-Schiphol line via Leiden. 
The case study uses confidential infrastructure data provided by the Dutch railway company ProRail.
The Appendix reports on the constructed network~$\network$ from this data, the details of the flexibility computation, and an analysis of the runtime components.
We used four scenarios based on real-world events; the details are also reported in the Appendix. 
For each scenario, we ran FlexSIPP to identify the tipping points for each train, as planned for the Eurostar. 
Table~\ref{tab:tip} shows the tipping points found for one of the four scenarios. 
Each row represents a regular train that is delayed by the time specified in the `Delay' column to accommodate the new Eurostar train passing at the point indicated by `Location'. 
The trains are numbered by their index in the scenario, and the delay location provides the city where the train must be delayed.
As the railway industry does not define any tipping points for this Eurostar scenario, we cannot make a direct comparison.
However, only two trains need to be slightly delayed to accommodate the Eurostar in this scenario.
So, for~\ref{q:tip}: yes, FlexSIPP provides tipping points for real-world cases with many trains in a congested network.

\begin{table}[t]
\centering
\caption{Tipping points found by FlexSIPP for Scenario~1(93 trains over a 237-minute timespan). Times in \texttt{mm:ss}.}
\label{tab:tip}
\begin{tabular}{llll}
\toprule
Tipping Point & Train & Location & Delay \\
\midrule
13:42 & 77 & Delft & 01:12 \\
22:20 & 58 & The Hague & 01:08 \\
\bottomrule
\end{tabular}
\end{table}

\begin{table}[b]
\centering
\caption{Any-start-time plans for Eurostar, showing the number of paths found for different starting times~$\earliestStartTime$, with the associated latest start time: the max~$\unsafeTime$ of those path ATFs.}
\label{tab:compare}
\begin{tabular}{cccccc}
\toprule
 &  & \multicolumn{2}{c}{\textbf{FlexSIPP}} & \multicolumn{2}{c}{\textbf{@MAEDeR}} \\
Scen & $\earliestStartTime$ & \# Paths & $\unsafeTime$ & \# Paths & $\unsafeTime$ \\
\midrule
\multirow[t]{3}{*}{1} & 2741 & 14 & 1507.80 & 11 & 1507.80 \\
 & 2799 & 1 & 2000.60 & — & - \\
 & 2934 & — & - & 1 & 2000.60 \\
\cmidrule{2-6}
\multirow[t]{2}{*}{2} & 2698 & 6 & 1244.30 & 3 & 1106.10 \\
 & 2743 & 1 & 2228.80 & 1 & 2228.80 \\
\cmidrule{2-6}
\multirow[t]{2}{*}{3} & 2328 & 3 & 1774.80 & — & - \\
 & 2373 & 2 & 1871.70 & 5 & 1871.70 \\
\cmidrule{2-6}
\multirow[t]{2}{*}{4} & 2334 & 1 & 1865.20 & — & - \\
 & 2335 & 12 & 1633.10 & 10 & 1888.90 \\
\bottomrule
\end{tabular}
\end{table}

We also compared the number of paths in the any-start-time plan found by @MAEDeR and FlexSIPP, as shown in Table~\ref{tab:compare}. 
This shows the start time when a path becomes available~$\earliestStartTime$ as well as the end of the availability interval~$\unsafeTime$.
As explained before, when the path requires waiting,~$\unsafeTime < \earliestStartTime$ can happen, which is the case in all the Eurostar paths.
In the first scenario, FlexSIPP finds the same path as @MAEDeR, though it has an earlier start time~$\earliestStartTime$ but the same end time~$\unsafeTime$, in other words, the same path is available earlier when using flexibility. 
FlexSIPP generally finds more paths than @MAEDeR, so the railway operator has more options to reroute the Eurostar.
In the third scenario, FlexSIPP found a smaller number of better plans that together dominate the set of plans @MAEDeR found by using flexibility.
So, FlexSIPP's any-start-time plans provide more flexible options to reroute the Eurostar through regular traffic, yielding more information, so we positively answer~\ref{q:info}.

\subsection{MovingAI Benchmarks}
FlexSIPP is most useful in contexts where paths are constrained, but agents have some buffer time along the paths.
We demonstrated this in the case study and now also show that FlexSIPP works on the regular MovingAI MAPF benchmarks \cite{Stern2019}.
Therefore, we chose two specific maps with all corridors of width~1: the maze of 128x128 with scenarios of~20 and~50 agents, and the warehouse of~20x40x10 with scenarios of~50 agents.
We computed the agents' original plans using Priority-Based Search \cite{Ma_Harabor_Stuckey_Li_Koenig_2019}.
For the maze, this yielded 25 and 22 scenarios for 20 and 50 agents, respectively, and 10 for the warehouse.

\begin{table}[t]
    \caption{Total delay comparison for maze scenarios with 20 and 50 ($\numAgents$) agents and warehouse scenarios with 50 agents, where one agent is delayed, and extra flexibility (\IntroducedFlexibility{}) is added in the initial paths. Shows the average delay difference in percentage (FlexSIPP / @MAEDeR) and the percentage of runs when a path was found, along with the average search time over those for FlexSIPP (F) and @MAEDeR (M).}
    \label{tab:delays}
    \begin{tabular}{ccclcccc}
        \toprule
         &  &  & \textbf{Delay} & \multicolumn{2}{c}{\textbf{Time (s)}} & \multicolumn{2}{c}{\textbf{Path Found}} \\
        Map & $\numAgents$ & \IntroducedFlexibility{} &  F / M & F & M & F & M \\
        \midrule
        \multirow[t]{4}{*}{\shortstack[l]{maze}} & \multirow[t]{4}{*}{20} & 0 & 93.4\% & 0.96 & 0.55 & 66\% & 64\% \\
         &  & 3 & 93.9\% & 1.51 & 0.81 & 66\% & 60\% \\
         &  & 5 & 93.8\% & 1.06 & 0.57 & 73\% & 70\% \\
         &  & 8 & 59.9\% & 2.02 & 0.80 & 64\% & 57\% \\
        \cmidrule{2-8}
         & \multirow[t]{4}{*}{50} & 0 & 91.8\% & 2.74 & 1.51 & 68\% & 64\% \\
         &  & 3 & 86.1\% & 3.78 & 1.77 & 66\% & 61\% \\
         &  & 5 & 93.0\% & 3.31 & 1.92 & 57\% & 52\% \\
         &  & 8 & 66.2\% & 3.54 & 1.87 & 70\% & 61\% \\
        \cmidrule{1-8} \cmidrule{2-8}
        \multirow[c]{4}{*}{\shortstack[l]{ware\\house}} & \multirow[t]{4}{*}{50} & 0 & 100.0\% & 3.56 & 3.26 & 70\% & 70\% \\
         &  & 3 & 100.0\% & 4.89 & 4.13 & 70\% & 70\% \\
         &  & 5 & 98.1\% & 4.69 & 4.15 & 53\% & 53\% \\
         &  & 8 & 102.5\% & 3.20 & 2.52 & 63\% & 63\% \\
        \bottomrule
    \end{tabular}
\end{table}

First, we considered the optimal path for the delayed agent with a delayed start time, based on the total delay of all agents, computed by summing the arrival time minus the original arrival time before the delay was handled.
To represent robustness and evaluate the effect of flexibility, we modified the plans while maintaining the paths of the original plan, so that whenever two agents use the same node within~\IntroducedFlexibility{} timesteps, the latter agent waits at the previous node until there is a difference of~\IntroducedFlexibility{} timesteps, where~$\IntroducedFlexibility{}=0$ is the original plan from Priority-Based Search. 
For each scenario, we randomly delayed one agent by selecting a location along its path and sampling a delay between 0 and the time the next agent enters that location. We repeated this three times per scenario to obtain different random delays.
The path returned by FlexSIPP was compared to the path returned by @MAEDeR, where the latter cannot delay other agents, whereas the former is optimal given that other agents can only be delayed, not rerouted. 
Table~\ref{tab:delays} shows the total delay for all agents that was found by FlexSIPP and @MAEDeR, respectively, and FlexSIPP was able to find paths in many more cases for the maze.
The difference (Diff) is averaged across scenarios and the three iterations per scenario; a negative value indicates that FlexSIPP's plan resulted in less overall delay.
For the warehouse, FlexSIPP does not outperform @MAEDeR because the agents interact much more frequently in their paths in this map, so they have less flexibility.
While FlexSIPP was a bit slower than @MAEDeR, we see that with additional flexibility in the original plans, FlexSIPP was able to find a better overall plan than @MAEDeR, answering~\ref{q:time}.
We also note that sometimes neither algorithm could find a new plan for the delayed agent, and full replanning for all agents would be required.

\begin{figure}
    \centering
    \includegraphics[width=.94\linewidth]{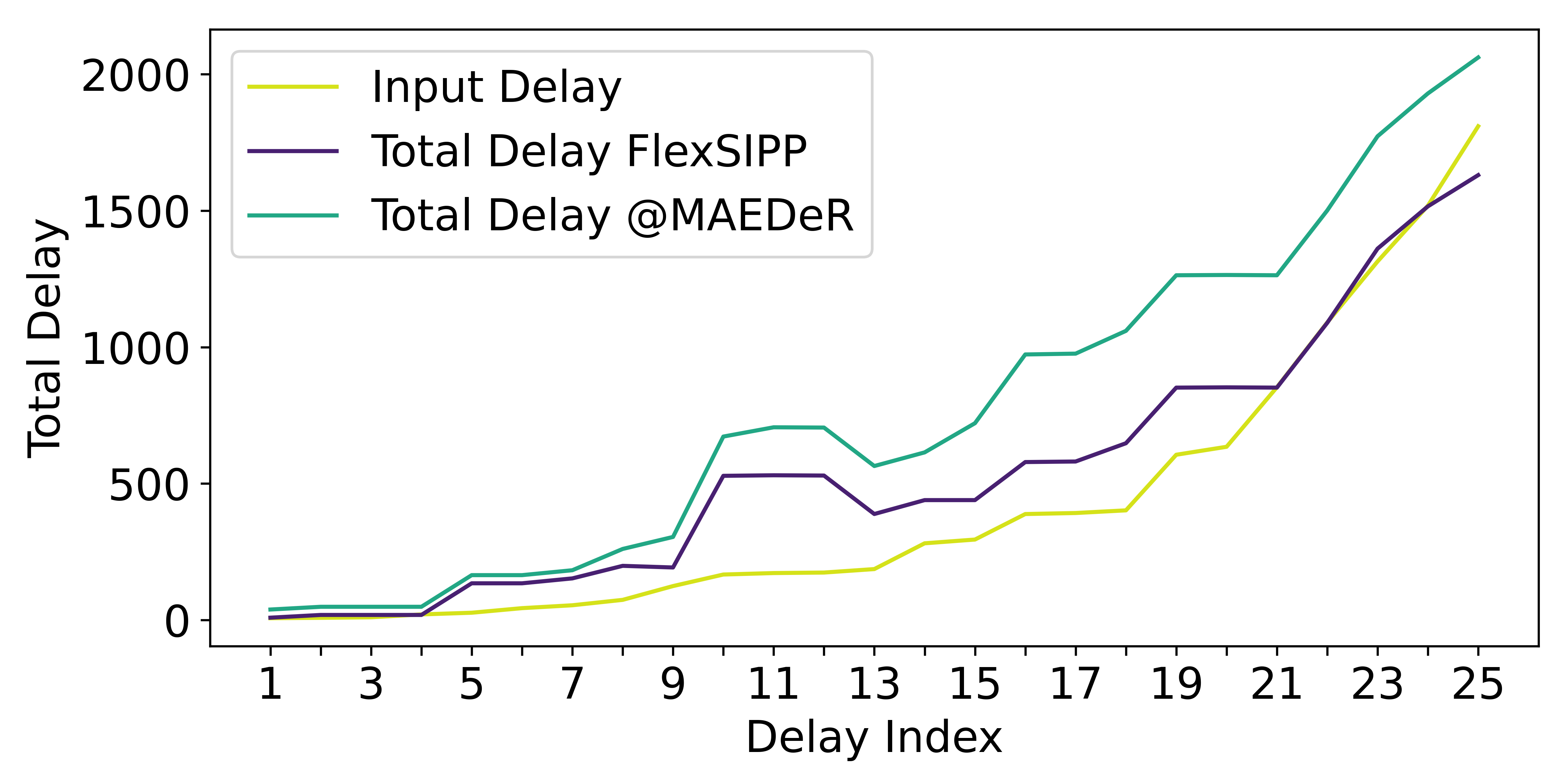}
    \caption{Total delay progression over sequential updates. The input delay is the agents' cumulative imposed delay, showing the delays in resulting arrival times. $\times$ marks runs where no path was found (and @SIPP graph not updated).}
    \label{fig:seq}
\end{figure}

Finally, we used the maze map with one scenario of 50 agents and their original paths ($\IntroducedFlexibility{}=0$).
Then, over 25 sequential iterations, we randomly delayed one agent.
FlexSIPP returned the fastest path for this delayed agent, after which we updated our @SIPP graph and handled the next delay. 
Figure~\ref{fig:seq} shows the difference in delays, where we take the input delay from the randomized delays and show the difference in original arrival times and resulting arrival times after each delay of FlexSIPP compared to @MAEDeR.
FlexSIPP always results in fewer delays and sometimes even yields a lower total delay than the input: if the original plan was not optimal or if a previously delayed agent is rerouted and a better plan becomes available. 
To answer~\ref{q:seq}: FlexSIPP can indeed plan with sequential delays.

\def\mathdefault#1{#1}

\section{Related Work}
\label{sec:rel}
The flexibility computation in Algorithm~\ref{alg:flex} is very similar to vertex slack \cite{Jiang2025}, although we define flexibility as the maximum amount we can delay an agent at a configuration without delaying \emph{any} other agent, whereas vertex slack is the maximum amount an agent can be delayed at a configuration without delaying another \emph{specific} agent.
Other order-switching algorithms were introduced before.
The Switchable \cite{Jiang2025} and Bidirectional \cite{Su2024} Temporal Plan Graphs define order switches between agents, repairing a plan by introducing delays. \citet{Kottinger2024} proposed to let an agent wait along its path, and the Location Dependency Graph \cite{Liu2024} avoids conflicts and deadlocks by coordinating the input wait and move actions.
However, neither method allows the originally delayed agent to be rerouted. 
Moreover, we provide a stronger guarantee on the impact (i.e., delay) of the original delay on the other agents. 
While the (Switchable) Action Dependency Graph \cite{Honig2019,Berndt2020} also seems related, it reconsiders the plans and orderings of all agents. 
In contrast, FlexSIPP focuses solely on replanning a single delayed agent when needed.
Contingency plans \cite{Nekvinda2021}, where the delayed agent can choose an alternative path, also precompute alternative routes, and collision-free paths are guaranteed, as only one agent can choose an alternative path.
Conversely, FlexSIPP reuses the precomputed @SIPP graphs, allowing it to quickly handle a new delay by updating the safe intervals of the newly delayed agent.

FlexSIPP relates to replanning for the Multi-Agent Pathfinding (MAPF) problem, in which agents must reach their goal locations without conflicts \cite{Stern2019}.
\citet{Svancara2019} used an extended problem formulation that allows new agents to appear and replans a different number of agents under various guarantees.
Two proposed methods do not allow changes to other agents' plans, as in MAEDeR's problem formulation, whereas the others impose fewer restrictions.
FlexSIPP strikes a new balance between computational efficiency and optimality by keeping the paths of all other agents but adjusting their delays to create an opening for the new agent to plan.
While we focus on recovering from a delay (using precomputation), \citet{Atzmon2018} proposed $k$-robust planning, providing plans that are robust to delays of upto~$k$ timesteps.

Our SIPP representation enables continuous-time solutions, related to Continuous MAPF approaches, such as Continuous Conflict-Based Search (CCBS) \cite{Andreychuk2019}, although it focuses on planning and does not allow replanning or flexibility.
In Lifelong MAPF, new goals are continuously added to the current scenario during execution. 
So, the plan must be continually updated to reflect the current situation. 
While different strategies for replanning are used, such as replanning all agents, or just the affected agents \cite{Zhang2024}, these methods do not reuse the knowledge from solving the scenario in earlier timesteps.
FlexSIPP could also be applied here to more efficiently use the information at different points in time and find a better overall solution.
FlexSIPP replans a single agent, which could iteratively build a multi-agent plan by adding one agent at a time, closely related to Prioritised Planning \cite{moragPrioritisedPlanningCompleteness2025}. 
However, deciding on the agents' priorities would also affect their flexibility.

\section{Conclusion}
\label{sec:conc}
FlexSIPP uses the available temporal flexibility in a multi-agent plan to precompute the any-start-time plans for all agents, so that upon a single-agent delay, this agent can immediately retrieve its safe paths, which may delay other agents, without propagating this delay further.
After such a delay, FlexSIPP handles new delays sequentially by updating the precomputed @SIPP graphs and returning the best solution for the single delayed start time.
The flexibility ensures no new conflicts arise with the delay, thus not further impacting any other agents.
So, we can reorder a delayed agent with other agents from the original plan. 
We demonstrated our method on a common MAPF benchmark and showed that it is effective in real-world settings, transforming detailed railway infrastructure into a simplified model via safe-interval path planning. 
While able to handle the continuous-time component and different agent speed profiles, our method is also effective in simpler grid-world settings, such as Multi-Agent Pathfinding.
Our method brings the single-agent replanning approach using any-start-time planning closer to a true multi-agent pathfinding approach.

\section*{Acknowledgements}
This work is part of the NWO LTP-ROBUST RAIL Lab, a collaboration between the Delft University of Technology, Utrecht University, NS, and ProRail. More information at \url{https://icai.ai/icai-labs/rail/}.

\bibliography{library-socs26}

\clearpage
\appendix
\section{Appendix: Railway Replanning with Flexibility}
\label{app:rail}

\begin{figure}[h]
    \centering
    \includegraphics[width=.95\linewidth]{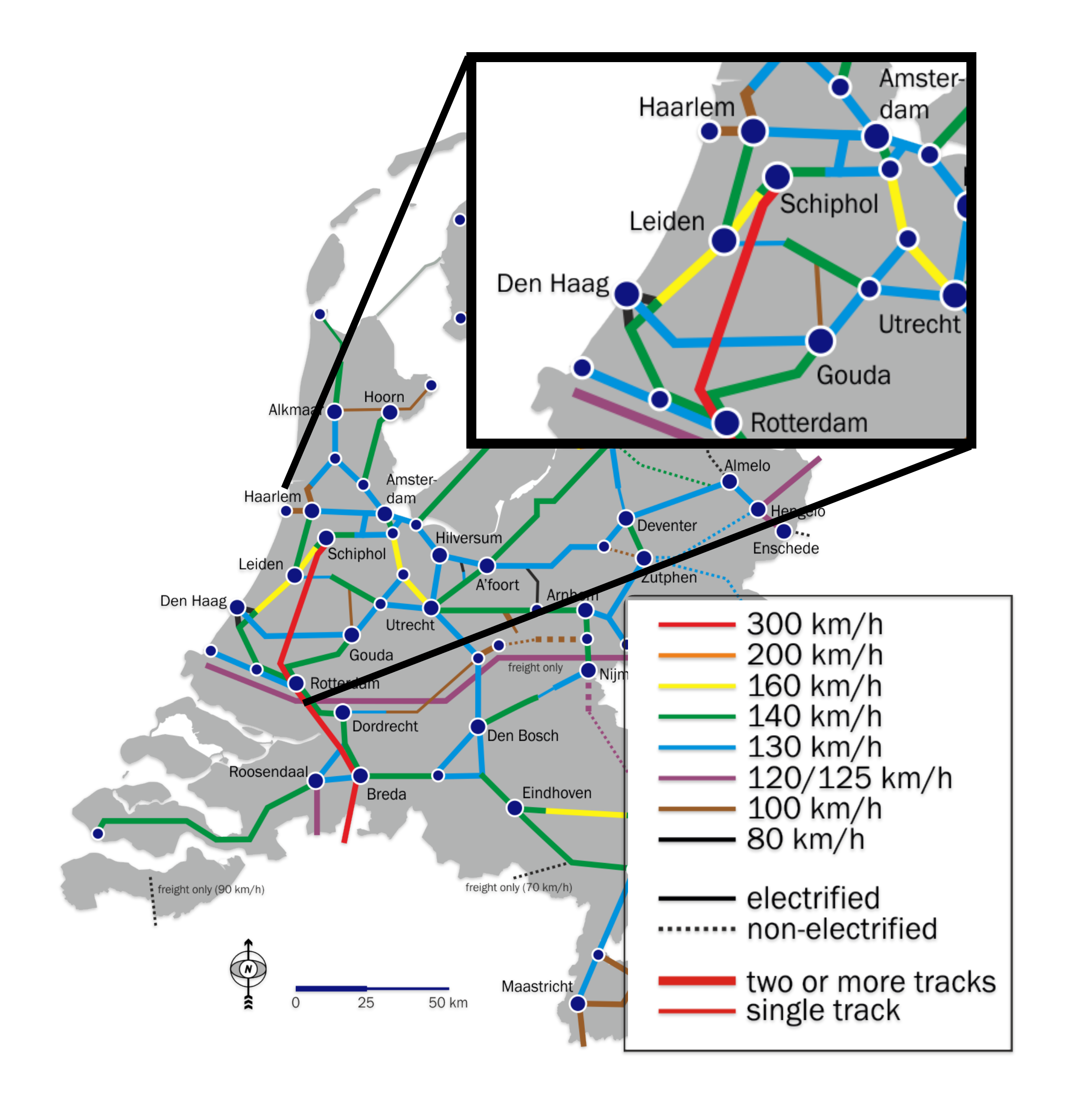}
    \caption{Railway Network in the Netherlands showing the maximum speed on each segment. Zoom-in of the area between Rotterdam and Amsterdam, both the regular track over Leiden in yellow, as well as the high-speed section between Rotterdam and Schiphol in red \cite{ProRail2022a}[p.199].}
    \label{fig:hsl}
\end{figure}

Figure~\ref{fig:hsl} shows the railway network in the Netherlands, with the section line between Schiphol and Rotterdam highlighted. 

The railway network is divided into \emph{blocks}, and only one train is allowed inside a block at a time. 
A common safety procedure uses a trackside signal to inform train drivers whether a block is occupied. 
To calculate when a block is free to enter, the railway industry uses \emph{blocking times}, which can be calculated precisely using the length of the block, the speed of the train, the reaction time of the driver, and the time it takes the train to clear the block \cite{Goverde2013}.
The blocking times can be visualized over the travel distance as in Figure~\ref{fig:head}.
In this example, two trains follow each other on the same route.
As long as the blocking times do not overlap, no conflict arises between these two trains.
We find the minimum separation of two trains, called the \emph{headway}, when two trains directly follow each other in at least one block, which is referred to as the critical block.
The flexibility of trains is thus easily visualized as the gaps between their blocking times, in industry referred to as~\emph{buffer time}, which also depends on the flexibility in the next block.

Now, we show how to model the Railway Replanning Problem as a~\problem{} problem.
Each block is a configuration~$\configuration$, allowing one agent to enter a block at a time.
We represent the railway network in a routing graph, connecting blocks based on whether a train travels directly from one to another \cite{Kemmeren2025}.
The blocking times are used to calculate the traversal times~$\transitTime$ between blocks.
From a given timetable, we generate the @SIPP graph for the delayed train and extract the trajectories of other trains, which are the dynamic obstacles.
Based on this timetable, we then compute the flexibility of the other trains, given by the blocking time computations.
Finally, we also allow the delayed agent to be a completely new agent that was not included in the original plan~$\sol$. 
In this case, the @SIPP graph that is given as input for~\problem{} has safe intervals that avoid conflicts with all agents from the original plan.

\begin{figure}
    \centering
    \begin{tikzpicture}[scale=0.45]
        \tikzset{
            rail/.style = {very thick},
            helper/.style = {thick, dotted},
            blockSlow/.style = {fill=black!10},
            blockFast/.style = {fill=black!40}
        }

        \def\fast{2.5}
        \def\block{3.5}
        \def\final{15.5}
        \def\diff{6}
        \def\start{1}
        \def\time{0.4}

        \draw[blockSlow](1,\start) rectangle (1+\block,\start+\fast);
        \draw[<->] (1.3,\start) -- node[right, pos=0.7, text width=3] {Blocking time} (1.3,\start+\fast);
        \draw[blockSlow](1+\block,\start+0.3*\fast) rectangle (1+2*\block,\start+1.6*\fast);
        \draw[<->] (1.2+\block,\start+0.3*\fast) -- node[right, pos=0.75, text width=3] {Blocking time} (1.2+\block,\start+1.6*\fast);
        \draw[blockSlow](1+2*\block,\start+0.9*\fast) rectangle (1+3*\block,\start+1.9*\fast);
        \draw[blockSlow](1+3*\block,\start+1.2*\fast) rectangle (1+4*\block,\start+2.6*\fast);

        \draw[blockFast](1,\start+1.7*\fast) rectangle (1+\block,\start+2.7*\fast);
        \draw[<->] (1.3,\start+1.7*\fast) -- node[right, pos=0.7, text width=3] {Blocking time} (1.3,\start+2.7*\fast);  
        \draw[blockFast](1+\block,\start+2*\fast) rectangle (1+2*\block,\start+3*\fast);
        \draw[<->] (1.2+\block,\start+2*\fast) -- node[right, pos=0.7, text width=3] {Blocking time} (1.2+\block,\start+3*\fast);  
        \draw[blockFast](1+2*\block,\start+2.3*\fast) rectangle (1+3*\block,\start+3.3*\fast);
        \draw[blockFast](1+3*\block,\start+2.6*\fast) rectangle (1+4*\block,\start+3.6*\fast);

        \draw[thick]  plot [smooth] coordinates { (0.5,\start+0.2) (1,\start+0.2*\fast) (1+\block,\start+0.6*\fast) (1+2*\block,\start+1.3*\fast) (1+3*\block,\start+1.6*\fast) (1+4*\block,\start+2.3*\fast) (\final, 1+2.4*\fast) } ;
        \node[left] at (0.5,\start+0.2) {Train 1};
        \draw[thick] plot [smooth] coordinates { (0.5,\start+1.8*\fast) (1+\block,\start+2.2*\fast) (1+2*\block,\start+2.6*\fast) (1+3*\block,\start+3*\fast+0.3) (\final, \start+3.5*\fast) };
        \node[left] at (0.5, \start+1.8*\fast) {Train 2};
        
        \draw[very thick, <->] (1+3*\block,\start+1.6*\fast) -- node[pos=0.75, right, text width=1.4cm] {\textbf{Headway}} (1+3*\block,\start+3*\fast+0.3);
        \draw[dashed, <->] (1+3*\block, \start) -- node[above, text width=1cm] {Critical block} (1+4*\block,\start);     
        
        \draw[rail] (0,0) -- (\final, 0);

        \foreach \xy in {(1, 0), (1+\block, 0), (1+2*\block, 0), (1+3*\block, 0), (1+4*\block, 0)}
        {   
            \path \xy ++(0, -0.15) coordinate (s1) -- \xy ++ (0, -0.45) coordinate (e1);
            \draw (s1) -- (e1);
            \path \xy ++(0, -0.3) coordinate (s1) -- \xy ++ (0.4, -0.3) coordinate (e1);
            \draw (s1) -- (e1);
            \draw \xy +(0.575, -0.3) circle (0.175);
            \path \xy coordinate (y1) -- \xy ++ (0, 10) coordinate (y2);
            \draw [helper] (y1) -- (y2);
        }
        \draw[->, thick] (-0.5, 7) -- (-0.5, 10) node[midway,sloped,above,align=center] {Time};
        \draw[->, thick] (6, -1.5) -- (9, -1.5) node[midway,sloped,above,align=center] {Distance};
            
    \end{tikzpicture}
    \caption{Two blocking time staircases of two trains along the same track. The rightmost block is the critical block. Adapted from \citet{Pachl2021}.}
    \label{fig:head}
\end{figure}

As a heuristic in the search procedure, we use the shortest distance to the goal configuration, without considering any other agents. 
This never overestimates the cost of reaching this destination, so the heuristic is \emph{admissible}. 
The heuristic is also \emph{consistent}, as it always decreases by at most the minimum edge duration when moving closer to the agent's goal location.

\subsection{Experiments}
From the infrastructure, the resulting routing graph of the entire Dutch railway network consists of 9700 nodes with 247,600 routes, averaging 24.5 outgoing routes per node, which is reduced in size to 2.48 outgoing nodes by discarding shunting yards \cite{Kemmeren2025}[p.35]. 
For the case study experiments, we use the timetable data provided by the Dutch Railways.\footnote{\url{https://www.ns.nl/reisinformatie/ns-api}} 
We gathered the data on July 8, 2025, constructing four scenarios: 
\begin{itemize}
	\item Scenario 1 has 93 trains in a 237-minute timespan (resulting in 3996 safe intervals with 29501 ATFs)
	\item Scenario 2 has 94 trains in a 258-minute timespan (resulting in 3929 safe intervals with 28509 ATFs)
	\item Scenario 3 has 85 trains in a 246-minute timespan (resulting in 3284 safe intervals with 24069 ATFs)
	\item Scenario 4 has 75 trains in a 264-minute timespan (resulting in 3157 safe intervals with 23784 ATFs)
\end{itemize}

\subsection{Runtime Results}
Table~\ref{tab:time} shows the runtime components for the four scenarios, comparing FlexSIPP and @MAEDeR.
The `Creation routes~$\network$' generates the routing graph from the railway infrastructure, then unsafe intervals (the conflicts) are generated, including the blocking time calculations, from which the flexibility is computed, after which they are converted to safe intervals. The last two rows give the search time of the different algorithms, as the first steps are the same for all three algorithms.
FlexSIPP spends more time searching than @MAEDeR, because it finds many more paths and thus needs to restart the search more often.
The other time components are equal across both methods, as they use the same construction of the @SIPP graph from the underlying data, except for the flexibility generation, which is skipped in @MAEDeR.
As the scenarios use the entire Dutch railway network as an underlying graph, the generation process takes quite some time.
However, the computations done by FlexSIPP can also be precomputed, similar to @MAEDeR.
So, while FlexSIPP does require more runtime to search through the graph, as this is significantly larger, this is reasonable when we can precompute this for the entire railway network in the Netherlands.

\begin{table}
    \caption{Table reporting the components of the runtime in seconds of the two algorithms for four different scenarios of replanning the Eurostar.}
    \label{tab:time}
    \begin{tabular}{lrrrr}
        \toprule
        Time Components & S1 & S2 & S3 & S4 \\
        \midrule
        Creation Routes~$\network$ & 47.08 & 48.65 & 46.14 & 46.18 \\
        Conflict gen. & 58532 & 49014 & 47730 & 48806 \\
        Flexibility gen. & 0.18 & 0.19 & 0.16 & 0.16 \\
        Interval gen. & 5.92 & 8.94 & 6.35 & 7.64 \\
        Search FlexSIPP & 129.68 & 146.62 & 143.15 & 77.76 \\
        Search @MAEDeR & 86.67 & 97.18 & 41.81 & 50.88 \\
        \bottomrule
    \end{tabular}
\end{table}

\end{document}